\documentclass[conference]{IEEEtran}
\IEEEoverridecommandlockouts
\usepackage{cite}
\usepackage{amsmath,amssymb,amsfonts}
\usepackage{algorithmic}
\usepackage{graphicx}
\usepackage{textcomp}
\usepackage{xcolor}
\usepackage{color, colortbl}
\usepackage[colorlinks,urlcolor=blue]{hyperref}
\def\BibTeX{{\rm B\kern-.05em{\sc i\kern-.025em b}\kern-.08em
    T\kern-.1667em\lower.7ex\hbox{E}\kern-.125emX}}
\usepackage{tabularx}
\usepackage{multirow}
\usepackage{bm}
\usepackage{amssymb}
\usepackage{pifont}
\newcommand{\xmark}{\ding{55}}%

\begin{document}
\title{Multi-Class 3D Object Detection Within Volumetric 3D Computed Tomography Baggage Security Screening Imagery}

\author{\IEEEauthorblockN{Qian Wang}
\IEEEauthorblockA{\textit{Department of Computer Science} \\
\textit{Durham University}\\
Durham, UK \\
}
\and
\IEEEauthorblockN{Neelanjan Bhowmik}
\IEEEauthorblockA{\textit{Department of Computer Science} \\
\textit{Durham University}\\
Durham, UK \\
}
\and
\IEEEauthorblockN{Toby P. Breckon}
\IEEEauthorblockA{\textit{Department of \{Computer Science $|$ Engineering\}} \\
\textit{Durham University}\\
Durham, UK \\
}
}

\maketitle
\begin{abstract}
Automatic detection of prohibited objects within passenger baggage is important for aviation security. X-ray Computed Tomography (CT) based 3D imaging is widely used in airports for aviation security screening whilst prior work on automatic prohibited item detection focus primarily on 2D X-ray imagery. These works have proven the possibility of extending deep convolutional neural networks (CNN) based automatic prohibited item detection from 2D X-ray imagery to volumetric 3D CT baggage security screening imagery. However, previous work on 3D object detection in baggage security screening imagery focused on the detection of one specific type of objects (e.g., either {\it bottles} or {\it handguns}). As a result, multiple models are needed if more than one type of prohibited item is required to be detected in practice. In this paper, we consider the detection of multiple object categories of interest using one unified framework. To this end, we formulate a more challenging multi-class 3D object detection problem within 3D CT imagery and propose a viable solution (3D RetinaNet) to tackle this problem. To enhance the performance of detection we investigate a variety of strategies including data augmentation and varying backbone networks. Experimentation carried out to provide both quantitative and qualitative evaluations of the proposed approach to multi-class 3D object detection within 3D CT baggage security screening imagery. Experimental results demonstrate the combination of the 3D RetinaNet and a series of favorable strategies can achieve a mean Average Precision (mAP) of 65.3\% over five object classes (i.e. {\it bottles, handguns, binoculars, glock frames, iPods}). The overall performance is affected by the poor performance on {\it glock frames} and {\it iPods} due to the lack of data and their resemblance with the baggage clutter.
\end{abstract}


\section{Introduction}
X-ray baggage security screening is widely used to maintain aviation security. Currently, multi-view X-ray is predominantly used in aviation security for cabin baggage screening. This traditional baggage screening process, using 2D X-ray scanners, has the disadvantage of both inter-object occlusion and clutter within any given image projection of the scanned baggage item. As a result, it poses a considerably challenging visual search task for the human operators to discover the prohibited items (e.g., liquids, firearms, knives, etc.) overlapped with other benign items (e.g., electronic devices) within a constrained time frame. For this reason, passengers are currently required to divest large electronic devices and liquids which decreases checkpoint throughput significantly. Furthermore, human operator performance can be subjective and is heavily affected by many factors such as the experience, fatigue, monotony and concentration, although many successful measures have been taken to alleviate the problem in practice (e.g., Threat Image Projection (TIP) \cite{bhowmik2019good,wang2020reference} and shorter shift rotations \cite{meuter2016and}).

By leveraging recent advances in object classification and detection, significant progress has been made in automatic prohibited item detection within 2D X-ray imagery \cite{akcay2018using}. The use of deep learning techniques allows real-time and accurate detection of prohibited items even in cluttered X-ray images \cite{gaus2019evaluating,gaus19firearms,bhowmik19subcomponent,bhowmik19electronics}. However, performance can be affected when the baggage contains significant clutter and inter-object occlusion due to the fundamental limitation of projected 2D X-ray imagery.
To improve the detection rate without affecting the checkpoint throughput, airports are currently increasing the use of 3D CT screening which does not require the removal of electronic devices and liquids during baggage screening. The reconstructed 3D CT images provide more information and make it possible for the human operators to inspect the 3D CT images from differing views. However, current technology does not facilitate the automatic detection of (non-explosive) prohibited items such as prohibited items and liquid containers. 
In the prior work \cite{wang2020evaluation}, it has shown the possibility of using deep 3D CNN models for object classification and detection within baggage security imagery.
However, the study was limited to the detection of only one specific object category (i.e. either bottles or handguns) by one model. It is unknown \textit{how the detection performance will be affected for multi-class object detection in a unified framework} and \textit{which strategies are beneficial to the enhancement of detection performance in 3D CT baggage security screening imagery?}

To answer the above questions, in this paper we extend the prior work \cite{wang2020evaluation} in single-class object detection to a more challenging multi-class object detection problem in a unified framework and propose a viable solution to this problem. The proposed approach is evaluated in real volumetric 3D CT baggage security screening imagery to get insightful observations and conclusions for this emerging research topic. Specifically, we investigate different CNN architectures (i.e. ResNet \cite{he2016deep}) with variable depths under the RetinaNet object detection framework \cite{lin2017focal}. We also evaluate the effectiveness of data augmentation techniques including 3D volume flipping and rotation. 

The contributions of this work are summarized as follows:
\begin{itemize}
    \item[--] a unified framework using deep CNN models for multi-class prohibited item detection within volumetric 3D CT baggage imagery;
    \item[--] an evaluation of different 3D CNN models in the detection of prohibited items within volumetric 3D CT baggage imagery and the effect of data/feature augmentation.
\end{itemize}
\section{Related Work}\label{sec:related}
In this section, we briefly review existing works related to ours in \textit{3D object detection} and \textit{baggage security screening}.

\subsection{3D Object Detection}
In many real-world applications, the task of object detection is needed within 3D data modalities such as RGBD, point cloud, 3D Computed Tomography (CT) and 3D MRI. In this section, we aim to review the advances of 3D object detection using 3D Convolutional Neural Networks (CNN) in the applications of autonomous driving, medical image processing and beyond. We demonstrate the success of 3D object detection in these areas has inspired the application of 3D CNN in prohibited item detection within 3D baggage screening imagery in this work.
\subsubsection{3D Object Detection in Autonomous Driving}
Object detection is one of the core techniques enabling autonomous driving. 3D CNN models are widely used for this purpose using data from multiple sensors such as LiDAR \cite{maturana2015voxnet,zhou2018voxelnet} and RGB-Depth cameras \cite{qi2018frustum}. 

VoxelNet \cite{zhou2018voxelnet} is an end-to-end 3D object detector specially designed for LiDAR data. It consists of three modules: feature learning network (subdivide the point cloud into many subvolumes/voxels, feature engineering + fully connected neural network), convolutional middle layer (3D convolution applied to the stacked voxel feature volumes, each subvolume/voxel is a feature vector) and region proposal networks.
VoxNet \cite{maturana2015voxnet} in a more generic model being able to handle different types of 3D data including LiDAR point cloud, CAD and RGBD data.
Qi et al. \cite{qi2016volumetric} improved the performance of VoxNet by introducing the auxiliary subvolume supervision to alleviate the overfitting issue.

RGB-Depth data can also be processed using 3D CNN by firstly extracting proposals from 2D RGB images using a 2D object detector and transforming the proposals and corresponding depth information into 3D point clouds \cite{qi2018frustum}. The generated 3D point clouds can be further explored by 3D CNN models such as PointNet \cite{qi2017pointnet}.

One essential distinction of 3D object detection in autonomous driving and baggage security screening is that the objects of interest for autonomous driving have fixed sizes and orientations (e.g., vehicles, pedestrians, cyclists, etc.). This prior knowledge can be considered for bounding box proposal and reduce the false positives. For example, the 3D bounding boxes for pedestrians should have similar dimension ratios across different scenarios since a pedestrian must be standing rather than lying on the street. By contrast, such prior knowledge does not exist in baggage object detection and prohibited objects can have arbitrary sizes, orientations and locations in a baggage CT scanning.

\subsubsection{3D Detection in Medical Images}
3D detection has been applied in medical image processing for automatic early diagnosis and screening based on 3D CT and MRI imagery\cite{zhu2018deeplung,xie2019automated}. Hu et al. \cite{hu2018deep} reviewed recent works on medical image based cancer detection and diagnosis most of which have employed 3D CNN schemes for detection.  Monkam et al. \cite{monkam2019detection} reviewed the advancement of detection and classification of pulmonary nodules using 3D CNN in CT imagery. 3D CNN frameworks such as 3D U-Net and 3D DenseNet, 3D Faster R-CNN have been employed for nodule detection and the ensemble of multiple CNN models (e.g., checkpoints, varying input sizes, multiple CNN) is used to reduce the false positive.

One of the limitations of 3D CNN based object detection in medical imagery is the lack of sufficient training data. When compared with datasets in 2D imagery, 3D data are more difficult to collect, store and annotate and hence the existing datasets are relatively small for deep model learning. Chen et al. \cite{chen2019med3d} attempted to combine multiple 3D CT datasets to address the data sparsity issue and explored the capability of transfer learning to boost the object detection tasks in medical imagery. Although it has proved promising to transfer knowledge among different medical datasets and varying medical tasks, it is much more challenging to take advantage of these existing medical data in our study due to the big gap between these two domains.

From the reviews of existing works on 3D CNN based detection, 3D extensions of the off-the-shelf objection detection frameworks such as Faster R-CNN and RetinaNet generally outperform others \cite{monkam2019detection}. As a result, we extend the RetinaNet to 3D in our work for multi-class object detection within 3D CT baggage screening imagery. Our approach is built on the framework of medical object detection in \cite{jaeger2019retina} but has been adapted to the particular application of baggage security screening where the objects of interests can have arbitrary sizes, orientations and locations in the baggage CT volumes. 

\subsection{Baggage Security Screening}
Automatic object detection and recognition algorithms have been proposed and evaluated for baggage aviation security screening based on 2D X-ray images \cite{akcay2018using,gaus19firearms,bhowmik2019good}. The use of CNN architectures and object detection frameworks boosts the performance with a high detection rate and a low false positive rate. For instance, Gaus et al. \cite{gaus2019evaluation} evaluate the effectiveness of Faster R-CNN \cite{ren2015fasterrcnn}, Mask R-CNN \cite{he2017maskrcnn} and RetinaNet \cite{lin2017focal} in detecting six different objects (i.e. bottle, hairdryer, iron, toaster mobile and laptop) in 2D X-ray baggage images.

To enable automatic baggage screening using 3D CT imagery, a variety of studies have been carried out in recent years \cite{wiley2012automatic,flitton20123d,mouton20143d,jin2015joint,flitton2015object,mouton2015materials,mouton2015review,wang2019approach,wang2020reference}.

One research direction is object segmentation based on the material and morphological structure \cite{wiley2012automatic,mouton2015materials,wang2019approach}. Specifically, Mouton et al. \cite{mouton2015materials} propose a two-stage approach for object segmentation within 3D CT imagery. A CT volume is firstly coarsely segmented based on the voxel intensity ranges of pre-defined materials. Subsequently, a variety of shape descriptors are computed as features for the random forest classifier to determine a segment resulted from the first stage is good (containing only one object) or bad (containing multiple objects and hence need further segmentation). Wang et al. \cite{wang2019approach} studied the issue of object segmentation and classification in 3D CT imagery and focused mainly on the material characteristics without considering any specific prohibited item (e.g., firearm, knife, etc.). An approach to 3D segmentation is proposed based on recursive morphological operations and the Support Vector Machines (SVM) were employed for the classification of three types of materials.

\begin{figure*}
    \centering
    {\includegraphics[width=0.7\textwidth]{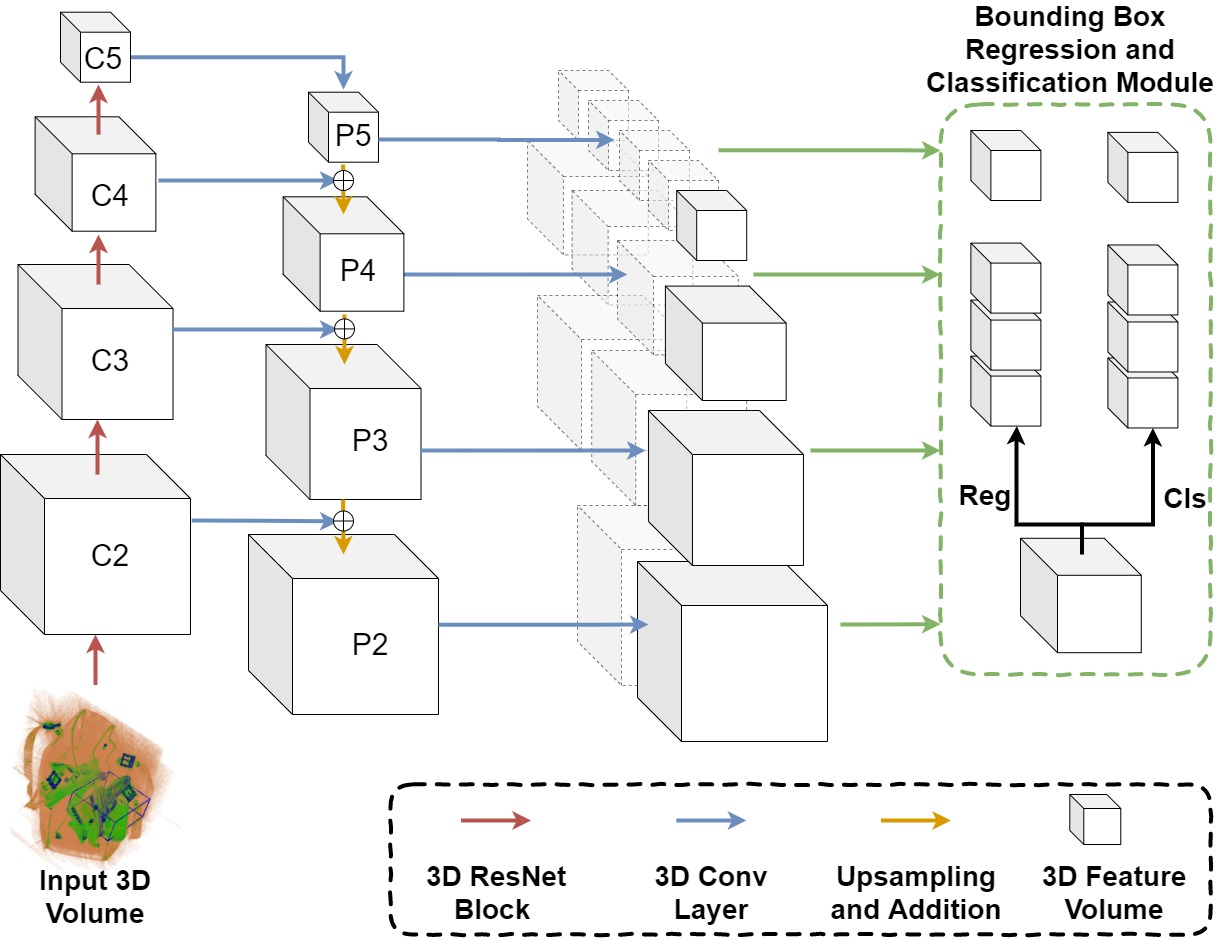}}
    {\caption{3D RetinaNet object detection framework with ResNet \cite{he2016deep} as the backbone model. }
        \label{fig:retinanet}}
\end{figure*}

3D object detection within 3D CT baggage security screening imagery has been studied in \cite{megherbi2010classifier,flitton2013comparison,wang2020evaluation}. Flitton et al. \cite{flitton2013comparison} evaluate the effectiveness of different 3D descriptors in a search-based detection approach. Their approach is limited to detect known objects for which the reference data are assumed to be available. Such an assumption hinders its application in practice when the reference data are usually unavailable. Wang et al. \cite{wang2020evaluation} use contemporary object detection frameworks based on 3D CNN and evaluate its performance on individual object detection independently. Based on this work, we present a unified framework for multi-class object detection within 3D CT imagery for baggage security screening.

\section{Method} \label{sec:method}
RetinaNet is employed in this study since it has proved better than its counterpart Faster R-CNN in the prior work \cite{wang2020evaluation}. We extend the RetinaNet framework to the 3D version used for 3D object detection in our study. Subsequently, data augmentation is described as a favourable technique to boost detection performance.

\subsection{3D RetinaNet} \label{sec:retinanet}
RetinaNet \cite{lin2017focal} is one of the most successful object detection frameworks designed for 2D natural images. Follow the same spirit, we extend it for 3D object detection within 3D CT imagery. As shown in Figure \ref{fig:retinanet}, our 3D RetinaNet consists of a feature pyramid network (FPN) implemented by ResNet \cite{he2016deep} and a 3D bounding box regression and classification module which is implemented by a few 3D convolutional layers.

The ResNet based FPN is formed by four ResNet blocks and the 3D feature volumes (corresponding to the feature maps in 2D) output from these four blocks are considered as \{C2, C3, C4, C5\} which have strides of \{4,8,16,32\} voxels with respect to the input volume. The top-down pathway and lateral connections are used to enhance the features generated in the bottom-up pathway (i.e. C2-5). The top-down feature volumes corresponding to C2-C5 are denoted as P2-P5. The highest-level feature volumes P5 are generated by a 3D convolutional layer with the stride of 1 and the kernel size of 3 from input C5. The feature volumes P4 are the summation of upsampled P5 and the output of a 3D convolutional layer with C4 as the input. Similarly, the feature volumes P3 and P2 are calculated.

Multi-scale feature volumes \{P2, P3, P4, P5\} are fed into the bounding box regression and classification module. The module consists of a branch for 3D bounding box regression and a branch for classification. These two branches have the same architecture with four 3D convolutional layers and an output layer. The output of the regression branch is a 3D volume with $6\times n_a$ channels corresponding to the bounding box biases with respect to the pre-defined anchor in a specific location. $n_a$ denotes the number of anchors pre-defined in each location. The output of the classification is a 3D volume with $6\times n_a$ channels corresponding to one background class and five foreground object classes under consideration in this study for each pre-defined anchor in a specific location.

The cross-entropy loss and smooth $L_1$ loss are used for the classification and regression respectively. The positive targets are calculated by comparing the pre-defined anchors against the ground truth bounding boxes with the Intersection Over Union (IOU) threshold of 0.1 which is also used as the threshold for detection during testing.

\subsection{Data Augmentation}\label{sec:method_det}
We investigate data augmentation for 3D volumetric CT data to enhance the object detection performance in our study. The 3D data augmentation strategies considered in this study are 3D threat image projection (TIP), data flipping and data rotation.

\subsubsection{Threat Image Projection}
\begin{figure}
    \centering
    {\includegraphics[width=0.5\textwidth]{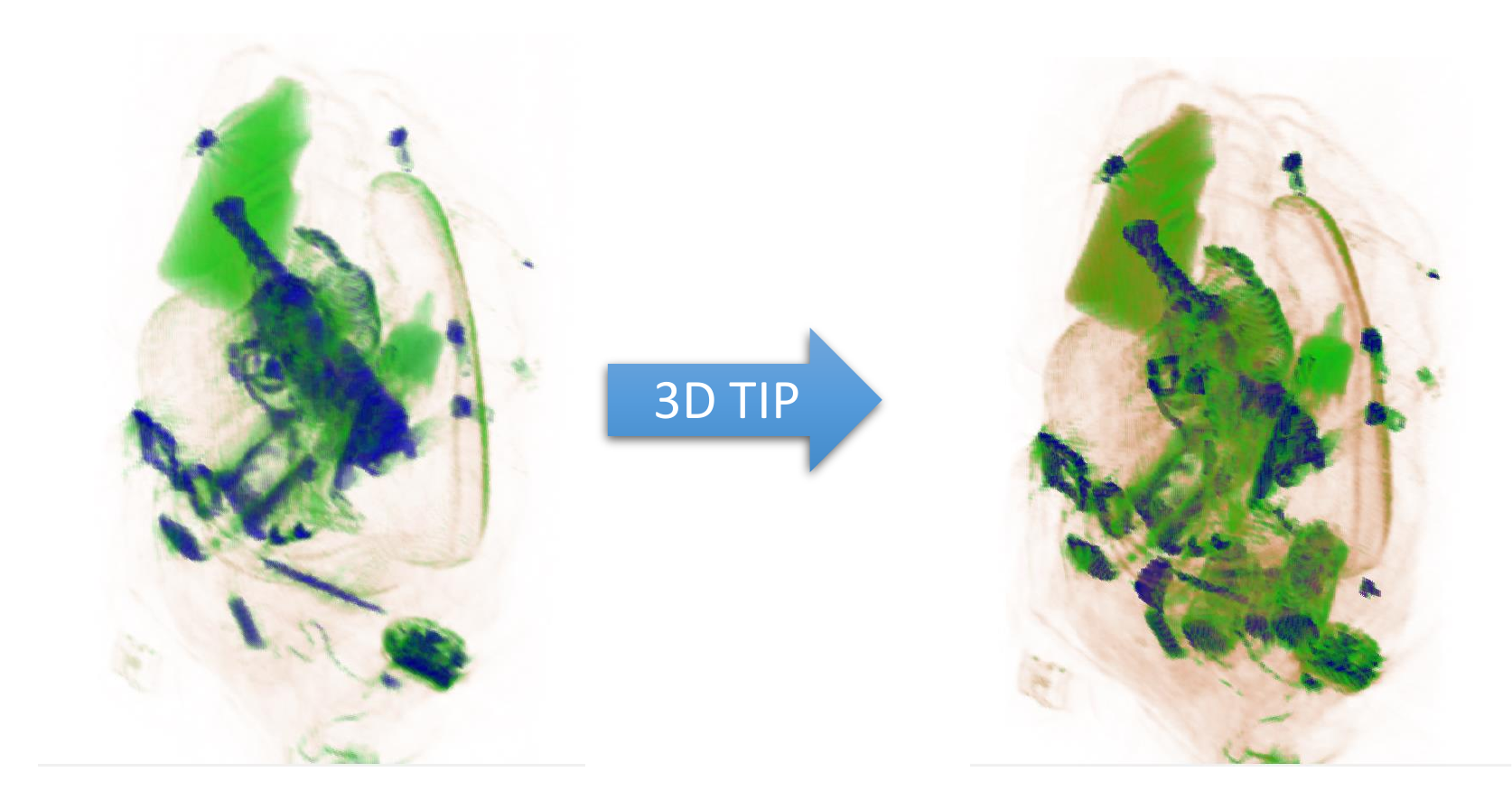}}
    {\caption{An illustration of 3D threat image projection for synthetic CT image generation (a signnature of \textit{binocular} is inserted into the bottom part of the baggage CT volume). }
        \label{fig:3dtip}}
\end{figure}

Threat image projection is a technique used in baggage security screening for training human screeners and automatic threat recognition algorithms \cite{bhowmik2019good}. Specifically, TIP approaches superimpose a threat item signature onto a benign baggage image to generate a realistic synthetic baggage image containing threat objects. Recently, the technique has been extended to 3D volumetric CT imagery \cite{wang2020reference}. We employ the approach presented in \cite{wang2020reference} to generate synthetic 3D volumes containing objects of interest. The isolated objects are first extracted from a CT volume and then inserted to other target CT volumes to generate more volumes with the objects of interest. 
As illustrated in Figure \ref{fig:3dtip}, we use 3D TIP techniques to insert a signature of \textit{binocular} into a baggage CT volume. We use this technique to address the issue of training data sparsity.

\subsubsection{3D Volume Flipping and Rotation}
To alleviate the overfitting issue in training, we employ volume flipping and rotation to augment training data randomly. For 3D volumetric CT data, the flipping can be conducted in three planes (i.e. \textit{x-y}, \textit{y-z} and \textit{x-z}). The rotation of a 3D volume is limited to 90 degrees around a specific axis which can be easily implemented by swapping axes. Each type of flipping and rotation ($6+6$) has a probability of $p$ to be activated during training so that the training data can be diversified significantly to alleviate the overfitting issue. We will investigate how the use of data augmentation and the value of $p$ can affect detection performance.

\section{Experimental Setup}\label{sec:experiments}
In this section, we describe the experimental setup for the evaluation of multi-class object detection within baggage CT volumes. We describe the dataset used in our experiments and implementation details of the detection methods.

\subsection{Dataset}
We create a dataset for experimental evaluation with data collected from a CT80-DR dual-energy baggage-CT scanner manufactured by Reveal Imaging Inc. Five object categories (i.e. {\it bottle, handgun, binocular, glock frame and iPod}) are considered in our experiments to simulate a multi-class 3D object detection problem. Due to the limited number of instances of {\it binocular, glock frame} and {\it iPod} (i.e. 16, 29 and 12 respectively) in the original CT volume data, we use the 3D Threat Image Projection (TIP) technique proposed in \cite{wang2020reference} to generate synthetically composited 3D CT volumes containing these object signatures of interest. As a result, the dataset is a combination of 478 real CT volumes and 287 synthetically composited ones generated by the TIP algorithm. The dataset is randomly divided into two subsets for training (70\%) and testing (30\%) respectively. Three random splits are used throughout our experiments. The detailed numbers of different object signatures within the dataset and three splits are shown in Table \ref{table:dataset}.

\begin{table}[ht]
\centering
\caption{Statistics of the dataset and data splits.}\label{table:dataset}
\begin{tabular}{lcccc}
\hline
 \multirow{2}{*}{Object} & Split 1 & Split 2  & Split 3  & \multirow{2}{*}{Total} \\
 &  (train/test) &  (train/test) & (train/test) \\ \hline \hline
\rowcolor{gray!20}
Bottle & 483/223 & 501/205 & 498/208 & 706\\ 
\rowcolor{gray!10}
Handgun & 269/108 & 263/114 & 267/110 & 377\\
\rowcolor{gray!20}
Binocular & 86/33 & 83/36 & 86/33 & 119\\
\rowcolor{gray!10}
Glock frame & 81/40 & 82/39 & 80/41 & 121\\
\rowcolor{gray!20}
iPod & 77/39 & 80/36 & 81/35 & 116 \\
\hline
\end{tabular}
\end{table}

\begin{table*}[ht]
\centering
\caption{Multi-Class Prohibited Object Detection Results (P: precision; R: recall; mAP: mean Average Precision).}
\label{table:resnet}
\newsavebox{\tablebox}
\begin{lrbox}{\tablebox}
\begin{tabularx}{1.14\textwidth}{c|cc|cc|cc|cc|cc|c}
\hline
\multirow{2}{*}{Model} & \multicolumn{2}{c|}{Bottle}& \multicolumn{2}{c|}{Handgun}& \multicolumn{2}{c|}{Binocular} &\multicolumn{2}{c|}{GlockFrame} &\multicolumn{2}{c|}{iPod} & \multirow{2}{*}{mAP (\%)} \\ \cline{2-11}
 &  P (\%) & R (\%) &  P (\%) & R (\%)&  P (\%) & R (\%)&P (\%) & R (\%)&P (\%) & R (\%)& \\ \hline \hline 
ResNet-10 & 80.2 $\pm$ 1.4 &70.9 $\pm$ 3.6& 77.5 $\pm$ 3.7 &83.1 $\pm$ 1.9& 75.4 $\pm$ 5.0 &84.3 $\pm$ 5.0& 72.3 $\pm$ 3.3 &44.1 $\pm$ 7.0& 29.4 $\pm$ 1.8 &37.6 $\pm$ 7.2& 58.2 $\pm$ 4.2 \\
ResNet-18 & 81.6 $\pm$ 3.6 &68.0 $\pm$ 4.0& 78.4 $\pm$ 0.6 &81.9 $\pm$ 1.1 &\bf 81.4 $\pm$ 3.4 &86.3 $\pm$ 8.7& 69.7 $\pm$ 8.9 &46.7 $\pm$ 7.0& 28.3 $\pm$ 3.8 &49.3 $\pm$ 6.1& 57.3 $\pm$ 4.4 \\
ResNet-34 &\bf 84.8 $\pm$ 2.8 &68.2 $\pm$ 2.9 &\bf 80.8 $\pm$ 7.4 &81.3 $\pm$ 1.2& 80.9 $\pm$ 2.0 &86.5 $\pm$ 7.9&\bf 78.5 $\pm$ 9.6 &36.5 $\pm$ 10.5& \bf 33.1 $\pm$ 4.4 &47.3 $\pm$ 1.9& 57.7 $\pm$ 3.8 \\
ResNet-50 & 81.0 $\pm$ 2.2 &\bf 74.0 $\pm$ 3.1& 74.8 $\pm$ 2.4 &\bf 91.0 $\pm$ 1.1& 78.4 $\pm$ 4.8 &\bf 87.3 $\pm$ 5.0& 63.9 $\pm$ 7.7 & \bf 53.5 $\pm$ 7.9& 32.8 $\pm$ 2.3 &\bf 56.6 $\pm$ 6.2& \bf 65.3 $\pm$ 3.6 \\
ResNet-101 & 80.6 $\pm$ 1.6 &71.1 $\pm$ 2.8& 71.3 $\pm$ 4.7 &89.8 $\pm$ 0.5& 75.5 $\pm$ 3.5 &85.1 $\pm$ 6.8& 70.6 $\pm$ 9.7 &51.8 $\pm$ 4.9& 30.2 $\pm$ 4.1 &53.6 $\pm$ 1.7& 62.8 $\pm$ 4.3 \\
\hline 
\end{tabularx}
\end{lrbox}
\scalebox{0.9}{\usebox{\tablebox}}
\end{table*}

\begin{table*}
\centering
\caption{Experimental Results of Multi-Class Object Detection with Different Data Augmentation Strategies.}\label{table:result_da}
\begin{tabular}{cc|ccccc|c}
\hline
 \multicolumn{2}{c|}{Data Augmentation} & \multicolumn{5}{c|}{Average Precision (\%)} & \multirow{2}{*}{mAP (\%)} \\ \cline{1-7}
Flipping & Rotation & Bottle & Handgun & Binocular & GlockFrame & iPod  \\  \hline \hline
 \xmark & \xmark & 65.3 $\pm$ 1.5& 79.7 $\pm$ 1.5& 69.2 $\pm$ 3.1& 32.5 $\pm$ 3.8& 28.2 $\pm$ 10.0& 55.0 $\pm$ 1.2 \\
 0.2 & \xmark & 67.5 $\pm$ 3.0& 84.5 $\pm$ 0.3& 81.5 $\pm$ 1.9& 48.3 $\pm$ 4.8& 37.6 $\pm$ 4.6& 63.9 $\pm$ 2.6 \\ 
 \xmark & 0.2 & \bf 70.1 $\pm$ 2.2& 77.8 $\pm$ 1.4& 74.9 $\pm$ 8.3& 43.0 $\pm$ 10.2& 28.1 $\pm$ 6.1& 58.8 $\pm$ 5.0 \\ 
 0.2 & 0.2 & \bf 70.0 $\pm$ 3.6& 84.9 $\pm$ 0.3& 83.7 $\pm$ 4.5& \bf 48.9 $\pm$ 7.1& \bf 38.9 $\pm$ 4.3& \bf 65.3 $\pm$ 3.6 \\  
 0.5 & 0.5 & 62.8 $\pm$ 2.1& \bf 85.6 $\pm$ 2.2& \bf 84.6 $\pm$ 3.0& 41.4 $\pm$ 8.7& 27.0 $\pm$ 8.5& 60.3 $\pm$ 2.6 \\
 
\hline
\end{tabular}
\end{table*}

\subsection{Implementation Detail}
The detection models evaluated in this work are implemented in PyTorch \cite{paszke2019pytorch} based on the work in \cite{jaeger2019retina}. In the experiments, we use the Adam \cite{kingma2015adam} optimiser with an initial learning rate of $1e-3$ for the first 100 epochs followed by a decreased learning rate of $1e-4$ for 100 epochs and $1e-5$ for last 100 epochs. This learning rate scheduler has been used throughout our experiments if not otherwise specified since it has been proved effective empirically in most cases.
All experiments are conducted on a GTX 1080Ti GPU.

\section{Experimental Results}\label{sec:results}
Thorough experiments are conducted to evaluate the effectiveness of the proposed approach to multi-class object detection in 3D CT baggage security screening imagery. Specifically, we evaluate varying ResNet \cite{he2016deep} architectures as the backbone FPN models, the effectiveness of data augmentation strategies, varying anchor sizes and scaling factors.

\subsection{On the Backbone Networks} \label{sec:res_backbone}
ResNet architectures \cite{he2016deep} with variant depths are employed as the backbones for FPN in the RetinaNet. We investigate the effect of different backbone models (i.e. ResNet10, ResNet18, ResNet34, ResNet50 and ResNet101) in this experiment. The experimental results are shown in Table \ref{table:resnet}. The precision and recall are reported for each object category with the mean and standard deviation over three splits. In addition, we also report the mean Average Precision (mAP) as the overall evaluation metric in Table \ref{table:resnet}. We can see ResNet50 performs the best overall with a mAP of 65.3\% over five object categories, followed by ResNet101 with a slightly lower mAP of 62.8\%. The other three architectures with less depth perform comparably with one another with the mAP around 57-58\%. Although ResNet50 achieves the best overall performance and the best recall rates, ResNet34 always results in better precision for all five object classes. By comparing the results of different object classes, we can see that \textit{Glock Frames} and \textit{iPods} have lower precision and recall than other three objects. This is due to the fact that {\it glock frames} are plastic hence more challenging to distinguish from background clutter within the baggage CT imagery whilst {\it iPods}, as a piece of electronic device, have less salient features to detect. In conclusion, the proposed approach, an extension of RetinaNet to 3D CT imagery, has the capability of detecting different objects within 3D CT baggage screening imagery but the performance varies across different object categories.

\subsection{On the Data Augmentation}
This experiment aims to investigate the effect of data augmentation strategies (i.e. volume flipping and rotation). It has shown using ResNet50 as the backbone network for FPN gives the best overall performance in the previous experiments, we use ResNet50 in this experiment. We compare the detection performance when no data augmentation is used and the performance when data augmentation is applied with different values of $p$ (i.e. 0.5 and 0.2).

The experimental results are presented in Table \ref{table:result_da}. When the data augmentation strategies are not used, a mAP of 55\% is achieved which can be boosted by the use of either random flipping or random rotation of the training data during training. The combination of two data augmentation strategies generates the best performance with the mAP of 65.3\% over five object categories. By increasing the probability of random flipping and rotation from 0.2 to 0.5, the overall performance degrades by a significant margin (as shown in the last row of Table \ref{table:result_da}). These results provide evidence that data augmentation is beneficial to the performance when properly used.

\begin{table*}
\centering
\caption{Experimental Results on volume scaling and anchor size.}\label{table:result_scaling}
\begin{tabular}{cc|ccccc|c}
\hline
 \multirow{2}{*}{Scaling factor} &\multirow{2}{*}{Anchor size} & \multicolumn{5}{c|}{Average Precision (\%)} & \multirow{2}{*}{mAP (\%)} \\ \cline{3-7}
 &  & Bottle & Handgun & Binocular & GlockFrame & iPod  \\  \hline \hline
 2 & 8-16-32-64 & 52.0 $\pm$ 1.5& 72.0 $\pm$ 4.8& 64.0 $\pm$ 4.3& 22.1 $\pm$ 7.0& 30.2 $\pm$ 6.2& 48.1 $\pm$ 4.3 \\
 3 & 4-8-16-32 & 61.0 $\pm$ 2.6& 81.7 $\pm$ 0.7& 77.0 $\pm$ 8.7& 32.8 $\pm$ 4.5& 19.3 $\pm$ 7.5& 54.4 $\pm$ 4.1 \\
 3 & 8-16-32-64 & \bf 70.0 $\pm$ 3.6& \bf 84.9 $\pm$ 0.3& \bf 83.7 $\pm$ 4.5& \bf 48.9 $\pm$ 7.1& \bf 38.9 $\pm$ 4.3& \bf 65.3 $\pm$ 3.6 \\  
 4 & 4-8-16-32 & 61.2 $\pm$ 0.4& 74.1 $\pm$ 2.3& 57.3 $\pm$ 8.2& 28.4 $\pm$ 2.6& 18.5 $\pm$ 7.8& 47.9 $\pm$ 1.9 \\
 4 & 8-16-32-64 & 64.2 $\pm$ 3.0& 81.7 $\pm$ 1.5& 76.1 $\pm$ 3.8& 36.3 $\pm$ 7.2& 10.6 $\pm$ 4.4& 53.8 $\pm$ 3.0 \\

\hline
\end{tabular}
\end{table*}

\begin{table*}
\centering
\caption{Experimental Results on High and Low Energy Data}\label{table:energy}
\begin{tabular}{c|ccc|ccc|c}
\hline
 \multirow{2}{*}{Data} & \multicolumn{3}{c|}{Bottle} & \multicolumn{3}{c|}{Handgun} & \multirow{2}{*}{mAP (\%)} \\ \cline{2-7}
 & Precision (\%) & Recall (\%)  & AP (\%) & Precision (\%) & Recall (\%)  & AP (\%)   \\  \hline \hline
 Low & 89.7 $\pm$ 2.1 &70.8 $\pm$ 2.6 &68.7 $\pm$ 2.7& 82.5 $\pm$ 1.5 &94.5 $\pm$ 2.4 &85.7 $\pm$ 3.7& 77.2 $\pm$ 3.0 \\
High & 90.4 $\pm$ 1.0 &68.5 $\pm$ 0.8 &66.3 $\pm$ 1.3& 83.6 $\pm$ 3.9 &92.6 $\pm$ 1.7 &85.7 $\pm$ 3.0& 76.0 $\pm$ 2.0 \\
High+Low & 90.2 $\pm$ 2.1 &71.0 $\pm$ 2.7 &68.9 $\pm$ 2.7& 82.9 $\pm$ 2.0 &92.9 $\pm$ 3.1 &84.0 $\pm$ 2.0& 76.4 $\pm$ 2.3 \\
\hline
\end{tabular}
\end{table*}

\begin{figure*}
    \centering
    {\includegraphics[width=0.95\textwidth]{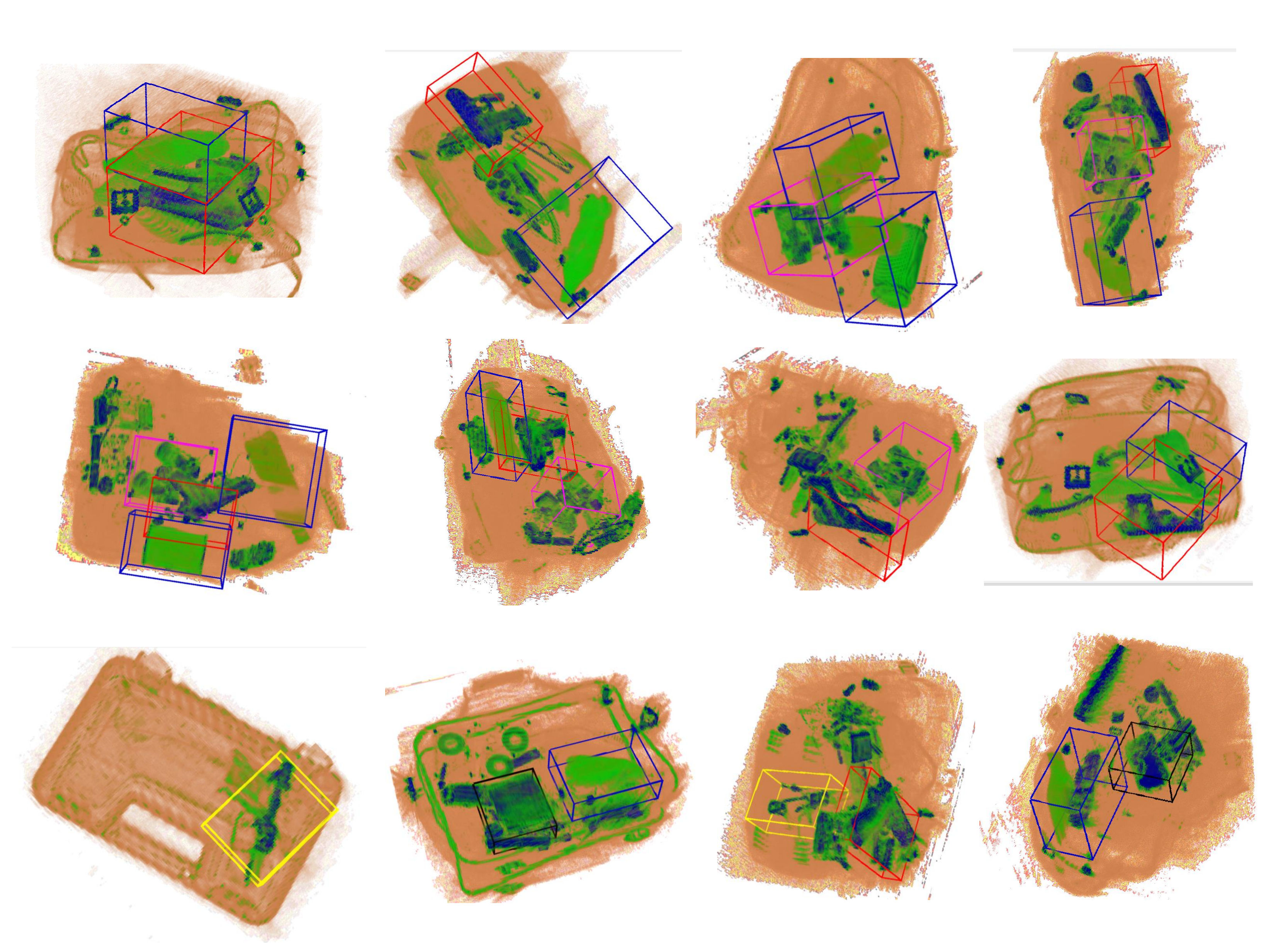}}
    {\caption{Detection results using the 3D RetinaNet with ResNet50 as the backbone network (the 3D bounding boxes for {\it bottles, handguns, binoculars, glockframes and iPods} are represented by blue, red, magenta, yellow and black colours respectively). }
        \label{fig:goodResults}}
\end{figure*}

\begin{figure}
    \centering
    {\includegraphics[width=0.5\textwidth]{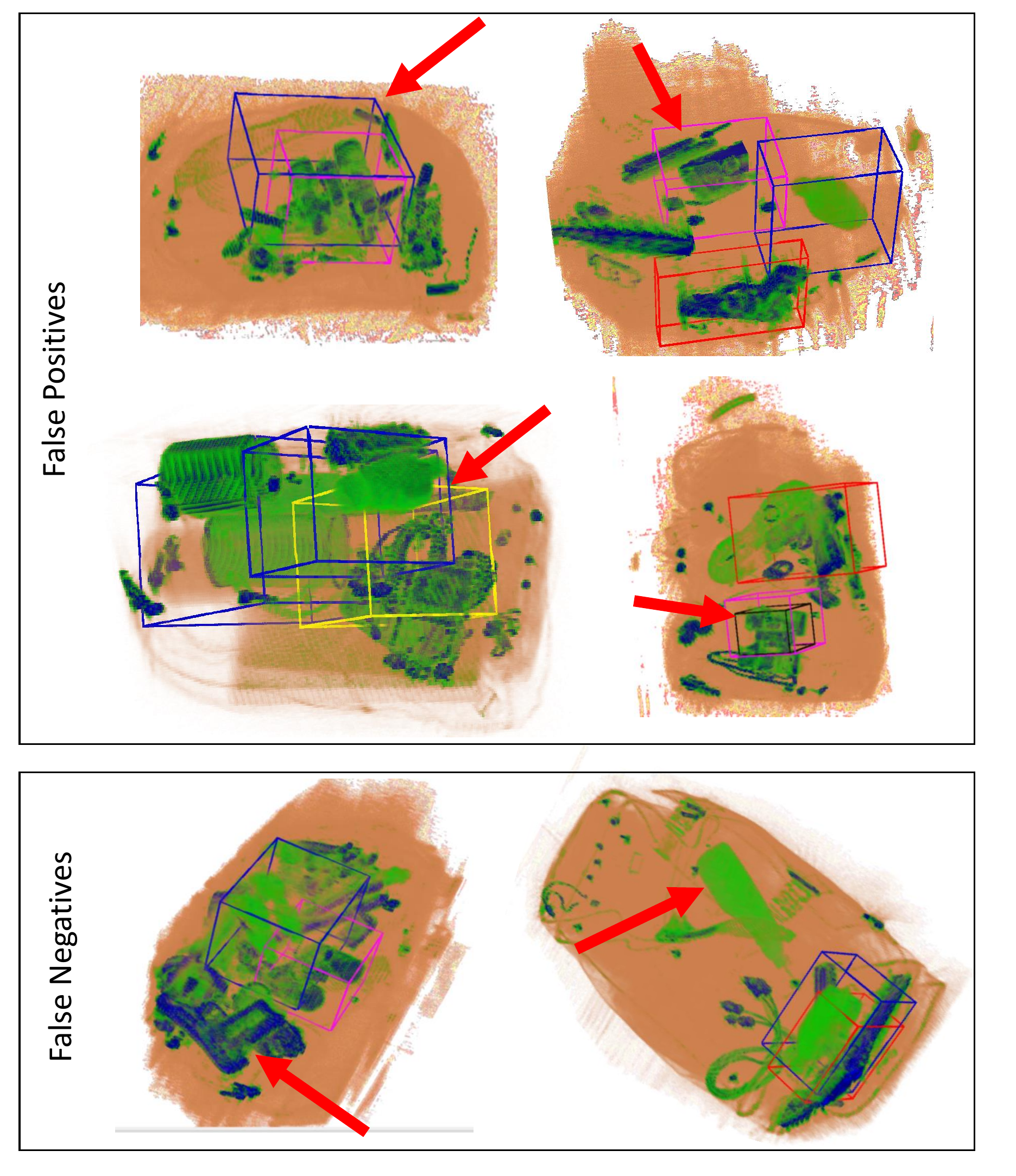}}
    {\caption{Exemplar false positive and false negative detection results (the false detection are emphasized with red arrows). }
        \label{fig:badResults}}
\end{figure}

\subsection{On the Anchor Size and Scaling Factor}
It is observed during the experiments that input CT volume scaling and anchor sizes have a significant effect on the performance of detection. We investigate how these two factors affect the detection results in this experiment. The scaling factor $s$ is a parameter used to down-sample the input CT volumes so that the down-sampled volumes will have $1/s$ of the original sizes in all three dimensions. The anchor size is the other crucial factor affecting the detection performance. Proper anchor sizes should be comparable with the ground truth bounding box sizes. It is easy to understand the anchor sizes should be adaptive to values of scaling factor $s$ for optimal detection performance. To these ends, in this experiment, we investigate different values of scaling factor $s$ as well as the anchor sizes.

The experimental results are displayed in Table \ref{table:result_scaling}. The best performance is achieved when the input CT volumes are down-scaled by a factor of 3 with an appropriate combination of anchor sizes for features in different pyramid levels. When the input CT volumes have higher or lower resolutions (i.e. scaling factor of 2 or 4), performance can be degraded significantly as shown in Table \ref{table:result_scaling} even the anchor sizes are adaptively adjusted. This observation may be caused by the fixed convolution kernel sizes and network architectures of the FPN which is worth further investigating in our future work.

\subsection{On the Raw Data}
We investigate how low- and high-energy raw data affect the object detection performance in this experiment. The CT volumes are reconstructed from raw CT slices generated by low or high energy X-ray. We also combine low and high energy data as two channels before feeding them into the 3D object detection networks. 
We use the optimal experimental settings derived from previous experimental findings and the results are shown in Table \ref{table:energy}. We use {\it Bottles} and {\it Handguns} as two representative prohibited items in this experiment since there are the most numbers of instances of them in the dataset (Table \ref{table:dataset}). It is demonstrated the low-energy and high-energy data lead to comparable object detection performance and the combination of them does not improve the performance. As a result, either low- or high-energy data from a dual-energy machine 
is necessary for the purpose of automatic object detection.

\subsection{Qualitative Evaluation}
To give qualitative evaluations of the proposed approach to 3D object detection within 3D CT baggage imagery, we present exemplar detection results in Figures \ref{fig:goodResults} - \ref{fig:badResults}. Figure \ref{fig:goodResults} list the detection results of eight typical CT volumes containing bottles, handguns or binoculars. The detected 3D bounding boxes are shown in different colours (i.e. blue, red, magenta, yellow and black for {\it bottles, handguns, binoculars, glock frames and iPods} respectively). The visualization in Figure \ref{fig:goodResults} demonstrates that the proposed approach can detect objects in varying orientations with relatively high localization accuracy. On the other hand, the approach also suffers from false positives and false negatives as shown in Figure \ref{fig:badResults}. False positives can be caused by misclassification of the target objects or non-target objects (i.e. background clutter having similar appearance characteristics to the target objects). False negatives can be caused by cluttered background or over-compact objects in the real baggage images. Overall, the poor detection performance of {\it glock frames} and {\it iPods} are caused by the high number of false positives and hence a low precision rate. It is also frequently observed that two predicted bounding boxes corresponding two different object categories (especially for binocular and iPods) overlap with each other with a high IOU value. This phenomenon can be caused by the artefacts introduced by 3D TIP in the synthetic CT volumes. Specifically, the artefacts rather than the real characteristics of the objects have been learned by the model for classification. This needs to be further investigated in future work with more real data available.

\section{Conclusion}
We address the multi-class object detection problem within volumetric 3D baggage security screening CT imagery. 3D RetinaNet is employed as the detector incorporated with different FPN architectures. 3D TIP and data augmentation techniques are employed to generate a synthetic dataset to alleviate the data sparsity issue. Experimental results validate the effectiveness of the proposed approach to the detection of five object categories in baggage CT volumes and also disclose the limitations of the current study (e.g., the lack of real data).

A few research directions following this work will be considered in our future work. Firstly, it is essential to scale up the dataset used for experiments in terms of both CT volumes and prohibited item types. Secondly, it is interesting to compare the effectiveness of 3D and 2D (slice based) CNN models in object detection within CT imagery. Finally, it is of great value to complement current approach by enabling the detection of material based prohibited items without specific shapes and appearances (e.g., explosive materials).
\ifCLASSOPTIONcaptionsoff
  \newpage
\fi

\bibliographystyle{IEEEtran}
\bibliography{ref}
\end{document}